\documentclass[runningheads]{llncs}
\usepackage{multirow}
\usepackage[T1]{fontenc}
\usepackage{amsmath}
\usepackage{amssymb}
\usepackage{array}
\usepackage{float}

\usepackage{graphicx}
\begin{document}
\title{Enhancing Image Classification in Small and Unbalanced Datasets through Synthetic Data Augmentation}
%
%
\author{Neil de la Fuente\inst{1,2} \and Mireia Majó \inst{1,2} \and Irina Luzko\inst{3} \and Henry Córdova\inst{3} \and Gloria Fernández-Esparrach\inst{3} \and Jorge Bernal\inst{1,2}}
\authorrunning{N. De La Fuente et al.}
%
\institute{Computer Vision Center, Barcelona, Spain \and
Universitat Autònoma de Barcelona, Barcelona, Spain\\ \and
Hospital Clinic, Barcelona, Spain\\
\email{\{ neil.delafuente , mireia.majo , jorge.bernal \} @ autonoma.cat }  \\
\email{\{ luzko, hcordova , mgfernan \} @ clinic.cat}}

%
%

%
\titlerunning{Enhancing Image Classification through Synthetic Data Augmentation}
%

\maketitle              

\begin{abstract}
Accurate and robust medical image classification is a challenging task, especially in application domains where available annotated datasets are small and present high imbalance between target classes. Considering that data acquisition is not always feasible, especially for underrepresented classes, our approach introduces a novel synthetic augmentation strategy using class-specific Variational Autoencoders (VAEs) and latent space
interpolation to improve discrimination capabilities.

By generating realistic, varied synthetic data that fills feature space gaps, we address issues of data scarcity and class imbalance. The method presented in this paper relies on the interpolation of latent representations within each class, thus enriching the training set and improving the model's generalizability and diagnostic accuracy.

The proposed strategy was tested in a small dataset of 321 images created to train and validate an automatic method for assessing the quality of cleanliness of esophagogastroduodenoscopy images.

By combining real and synthetic data, an increase of over 18\% in the accuracy of the most challenging underrepresented class was observed. The proposed strategy not only benefited the underrepresented class but also led to a general improvement in other metrics, including a 6\% increase in global accuracy and precision.

\keywords{Synthetic Data Augmentation  \and Variational Autoencoder \and Esophagogastroduodenoscopy Image Classification \and Image Classification}
\end{abstract}

\section{Introduction}

Gastric cancer (GC) is the $5^{\text{th}}$ most common cancer worldwide and there were more than 1 million new cases of GC reported in 2020. Esophagoduodenoscopy (EGD) is the gold standard method for the diagnosis of GC: several studies show that the detection of GC at earlier stages has a clear impact in the decrease of the mortality (hazard ratio [HR] 0.51) \cite{Ezoe2011,Guillena2019}. Nevertheless, up to 10\% of the cancers are missed during the exploration, with a clear impact on patient's survival rate \cite{Tsukuma2000}. Poor mucosal visualization is one of the factors that can negatively affect the diagnostic accuracy of gastric cancer.

For this reason, the degree of cleanliness and the quality of gastric mucosa visibility are of paramount importance. However, no broadly accepted cleanliness scale for the upper gastrointestinal tract (UGI) has been uniformly accepted and used in routine practice. Two scales have been recently published: POLPREP \cite{Romańczyk2022} and Barcelona scale \cite{Cordova2023}. Both evaluate the level of cleanliness in the esophagus, stomach and duodenum. They differ on the number of levels (4 for POLPREP, 3 for Barcelona scale) and in the degree of evaluation detail: Barcelona scale further divides stomach by segments (fundus, corpus and antrum).

However, these scales are prone to a certain degree of subjectivity. To cope with this, and following other methods already developed to assist clinicians in similar tasks \cite{Haithami2022}, there is room for AI systems that can provide an objective assessment of the degree of UGI cleanliness by an automatic classification of EGD images. The benefits of such a system are clear: if clinicians can be sure of those cases when gastroscopies are inappropriate due to insufficient cleanliness, they can make a recommendation to repeat the exploration. In the opposite case, where the UGI is clean, unnecessary repetitions can be avoided with the consequent saving of scarce economic resources.

The main technical challenge in medical image classification comes from the limited size and imbalance of available datasets. This limitation reflects the real-world shortage of annotated medical images and uneven distribution of pathological findings, making it difficult to develop robust models with traditional deep learning which usually requires large, balanced datasets. Additionally, the detailed nature of EGD images, which needs accurate identification of different levels of cleanliness, adds to the challenge. The scarcity of significant features in smaller datasets can result in model biases or underperformance. Overcoming these challenges requires innovative approaches that improve data diversity and representation, enhancing model's ability to generalize in real-world scenarios.

The key contributions of our work include:
\begin{itemize}
    \item \textbf{Use of class-Specific variational autoencoders (VAEs):} By generating synthetic images through latent representation interpolation within classes, we can expand the feature space and directly address class imbalance.
    \item \textbf{Focused enrichment of feature space:} Our technique fills gaps in the feature space with realistic synthetic images, improving training effectiveness and model sensitivity to critical subtle features for accurate classification.
    \item \textbf{Proof of the versatility of our approach across architectures:} We demonstrate the benefits of our methodology across two prominent image classification architectures such as EfficientNet-V2 \cite{EfficientNetV2} and ResNet-50.
\end{itemize}

\section{Related Work}

The previously mentioned scarcity of annotated medical datasets has led to novel strategies for data augmentation in image classification. 

Garay-Maestre et al. \cite{GarayMaestre2018} exploited Variational Autoencoders (VAEs) to generate synthetic samples, demonstrating how synthetic data, when combined with traditional augmentation methods, can improve the robustness and performance of machine learning models. Following this line of thought, Auzine et al. \cite{Auzine2022} applied Generative Adversarial Networks (GANs) with conventional augmentation to enhance the accuracy of deep learning architectures, such as ResNet50 \cite{ResNet} and VGG16 \cite{VGG16}, on endoscopic esophagus imagery. Further advancing the field, Zhou et al. \cite{Zhou2023} introduced 'Diffusion Inversion', a technique for creating synthetic data by manipulating the latent space of pre-trained diffusion models to achieve comprehensive data manifold coverage and improved generalization.

Liu et al. \cite{Liu2018} investigated data augmentation via latent space interpolation for image classification, showing significant improvements in model performance. Oring \cite{Oring2020} and Cristovao et al. \cite{Christovao2020} also explored the use of VAEs for generating in-between images through latent space interpolation, emphasizing the potential of this approach for enhancing data diversity. Moreno-Barea et al. \cite{MorenoBarea2020} and Elbattah et al. \cite{Elbattah2021} focused on improving classification accuracy using data augmentation techniques on small datasets, highlighting the effectiveness of synthetic data in addressing class imbalance. Wan et al. \cite{Wan2017} specifically addressed imbalanced learning through VAE-based synthetic data generation, demonstrating its potential to improve model performance on imbalanced datasets.

While data augmentation has been widely used in several research domains there is no work, to the best of our knowledge, that applies this technique to assess the degree of cleanliness of EGD images. Nevertheless there are works applied to similar images, such as the work of Nam et al. \cite{nam2021development}, which use InceptionResnetV2 to classify wireless capsule endoscopy images to determine the degree of mucosa visualization or the work of Zhu et al. \cite{zhu2019cnn}, which applies a compact convolutional neural network with 2 Densenet layers to label bowel preparation in colonoscopy images according to Boston Bowel Preparation Scale.

While some of the previously mentioned studies have demonstrated the potential of generative models for data augmentation, our approach specifically targets class imbalance in EGD image classification by leveraging class-specific VAEs. By generating realistic synthetic images for each class, we aim to fill gaps in the feature space, thereby enhancing the training process and improving sensitivity to subtle features. This synthetic data augmentation methodology stands out by directly addressing the challenges posed by small and unbalanced datasets, particularly for EGD images, and notably improving performance. 

\section{Methodology}

Variational Autoencoders (VAEs) \cite{KingmaWelling2014} represented a groundbreaking shift in the generation of synthetic data by providing a probabilistic approach to data encoding and decoding. A VAE is composed of an encoder, for translating input data into a latent space representation, and a decoder, for reconstructing data from this latent space. The encoder function, denoted as \(q_\phi(z|x)\), maps an input \(x\) to a latent space representation \(z\), parameterized by \(\phi\).

This process introduces a stochastic element by generating a distribution characterized by mean \(\mu\) and variance \(\sigma\), rather than a fixed point for the latent variables. This distribution allows for the sampling of new data points from the latent space, using the reparameterization trick: \(z = \mu + \sigma \odot \epsilon\), where \(\epsilon\) is an element-wise product with a random noise vector.

The decoder, denoted as \(p_\theta(x|z)\) and parameterized by \(\theta\), reconstructs the data from the latent space representation. The objective of training a VAE is to minimize the loss function \(\mathcal{L}(\theta, \phi; x)\), which is a combination of the negative log-likelihood of the reconstructed data and the Kullback-Leibler (KL) divergence, promoting an effective balance between data reconstruction fidelity and distribution approximation:

\begin{equation}
    \mathcal{L}(\theta, \phi; x) = -\mathbb{E}_{q_\phi(z|x)}[\log p_\theta(x|z)] + \text{KL}(q_\phi(z|x) \| p(z))
\end{equation}

The flexibility in generating new data points through this probabilistic framework makes VAEs particularly suitable for tasks like medical image augmentation, where capturing the diversity of pathological features is crucial.

The capabilities of VAEs are leveraged to generate synthetic images tailored for each class within the dataset. By training class-specific autoencoders, the unique characteristics and nuances of each class are captured in the latent space.

\begin{figure}[t!]
    \centering
    \includegraphics[width=1.\linewidth]{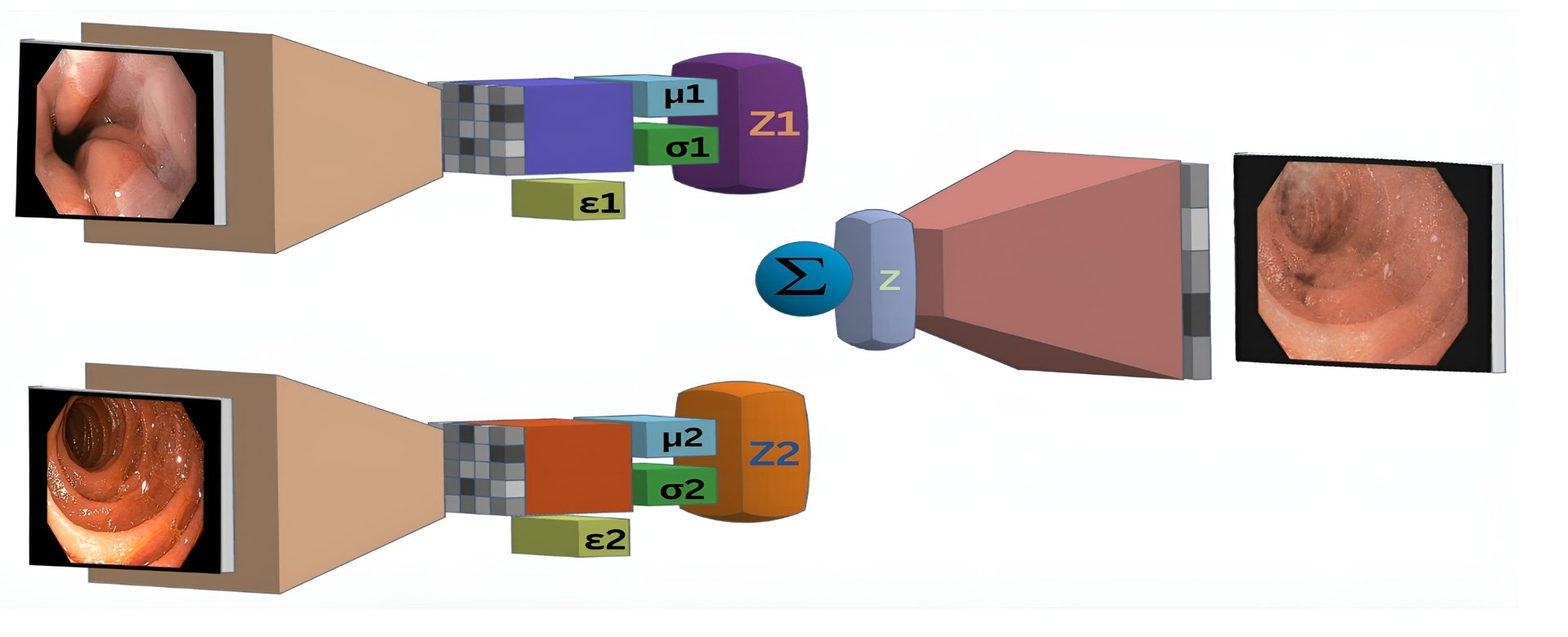}
    \caption{The encoding and decoding process in VAEs for synthetic image generation via latent vector interpolation.}
    \label{Synthetic data generation architecture}
\end{figure}

Post-training, the VAE synthesizes new images by performing an interpolation between the latent representations of two images within the same class, with the process illustrated in Figure \ref{Synthetic data generation architecture}. This interpolation is achieved by computing a weighted sum of their latent vectors \(z_1\) and \(z_2\), yielding a new latent representation, denoted as \(z_{interp}\):

\begin{equation}
z_{interp} = \alpha z_1 + (1 - \alpha) z_2 \
\end{equation}

\noindent where \(\alpha\) corresponds to the interpolation weight $[0,1]$. This parameter modulates the contribution of each original image to the synthesized image. The resulting latent vector \(z_{interp}\) is then decoded to generate a new synthetic image. This new image merges characteristics of the parent images while maintaining the class's defining features, effectively enriching the dataset's diversity and quantity.

By integrating additional synthetic images for each class, this iterative and monitored approach not only augments the dataset but also introduces meaningful and realistic variations which enrich the feature space.

The incorporation of these synthetic images into the training process is designed to enhance the classifiers' performance, particularly by improving its understanding of the underlying distribution of the underrepresented classes, thereby ensuring a more comprehensive representation of each class's feature space and subsequently elevating model accuracy.

\section{Experimental Setup}

\subsection{Dataset}

Our study utilized a dataset comprising 321 esophagogastroduodenoscopy (EGD) images, capturing various stomach regions including the esophagus, duodenum, antrum, body, and fundus. Images were labelled into different categories according to different degree of stomach cleanliness by seven clinicians following the definitions of the Barcelona scale. 

\begin{figure}[b!]
    \centering
    \includegraphics[width=1\linewidth]{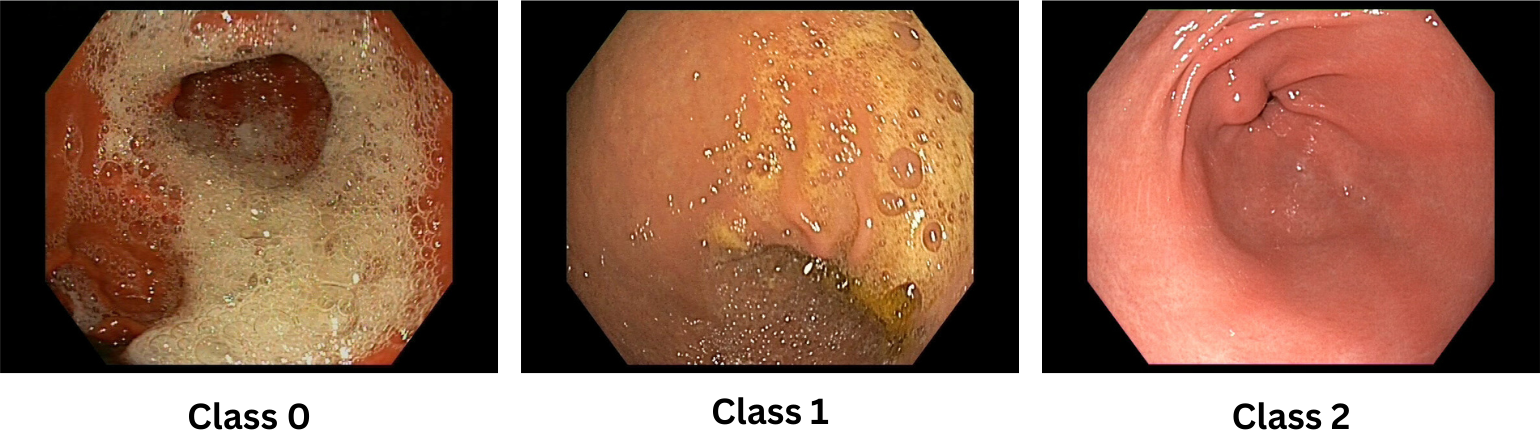}
    \caption{Sample image for each of the classes, labelled according to Barcelona scale.}
    \label{fig:Sample image for each of the classes}
\end{figure}

The consensus among the experts determined the final class assignment: 1) class 0, which corresponds to images with presence of non aspirable solid or semisolid particles, bile or foam preventing from clear mucosa visualization; 2) class 1, which corresponds to images with small amount of semisolid particles, bile or foam and 3) class 2, which comprises images without any kind of rest, allowing a complete visualization of the mucosa.  We show in Figure \ref{fig:Sample image for each of the classes} an example of some of the images in the dataset.

With respect to class distribution within the dataset, class 0 was represented with 65 images, class 1 with 91 and class 2 with 165. The dataset was split in a standard 80-20 fashion for training/validation and testing.

\subsection{Metrics}

To assess model performance, a suite of metrics was employed, including overall accuracy, precision, recall, and F1-score, complemented by class-specific measures for a thorough analysis. The evaluation concentrated on the efficacy of several augmentation strategies, encompassing traditional methods such as rotations and mirroring, as well the proposed approach involving synthetic data generation via VAEs. Results were tabulated to compare the impact of these techniques on model performance.

\subsection{Implementation Details}

\textbf{Synthetic Data Generation:} The synthetic data was generated using class-specific Variational Autoencoders (VAEs). Each VAE was trained separately for each class, capturing the unique characteristics of the respective class. The architecture of the VAE consisted of an encoder and decoder, with the encoder mapping input images to a latent space and the decoder reconstructing images from the latent space. We selected a latent space dimension of 256 based on empirical analysis, and the VAEs were trained for 1000 epochs with a learning rate of 0.0001, using the Adam optimizer. To generate synthetic images, we performed latent space interpolation between pairs of latent vectors within the same class, ensuring realistic and varied synthetic samples. We experimented with different quantities of synthetic images, ultimately finding that adding 300 extra images per class yielded the best performance, hence, results shown in section \ref{section:results} include 300 synthetic images per class.

\textbf{Classification:} For the classification task, we employed two well-known architectures: EfficientNet-V2 and ResNet-50. Both pretrained models were implemented and fine-tuned using PyTorch. The training process involved a batch size of 24 images and an initial learning rate set to \(5 \times 10^{-4}\). The learning rate was dynamically adjusted based on performance plateaus, with a reduction factor of 10 upon stagnation in validation loss improvement. To ensure reproducibility and consistency across experiments, a random seed was set to 42. 

Multiple configurations were explored for both EfficientNet-V2 and ResNet-50 architectures, including training on real data with and without traditional augmentations, and extending the dataset with synthetically generated images through VAE interpolation. Synthetic images were integrated into both classically augmented and non-augmented training sets to assess the combined effect of classical and synthetic augmentation techniques on the models' ability to generalize and accurately classify EGD images.

\section{Results}\label{section:results}

Experimental results, presented in Table \ref{tab:results}, remark the efficacy of incorporating synthetic data augmentation via class-specific VAEs in enhancing classification performance. A comparative analysis across different models and augmentation strategies reveals that the addition of VAE-generated synthetic images leads to substantial improvements in overall accuracy, precision, recall, and F1-scores.

\begin{table}[t!]
\centering
\caption{Classification performance of EfficientNet-V2 and ResNet-50 models with various data augmentation strategies.}
\resizebox{\textwidth}{!}{%
\begin{tabular}{|l|l|c|c|c|c|c|c|c|}
\hline
\textbf{Model} & \textbf{Data} & \textbf{Overall Acc.} & \textbf{Overall Prec.} & \textbf{Overall Rec.} & \textbf{Overall F1} & \textbf{Class-0 Acc.} & \textbf{Class-1 Acc.} & \textbf{Class-2 Acc.} \\ 
\hline
\multirow{3}{7em}{EfficientNet-V2} & Real No Aug. & 85.94 & 86.18 & 86.4 & 85.3 & 92.81 & 64.05 & 96.85 \\ 
& Real with Aug. & 87.5 & 87.51 & 87.5 & 87.46 & 81.43 & 74.56 & 97.10 \\
& Real + Gen No Aug. & 89.06 & 88.96 & 88.88 & 88.74 & \textbf{92.85} & 72.22 & 97.15 \\
& Real + Gen with Aug. & \textbf{92.19} & \textbf{92.27} & \textbf{91.03} & \textbf{91.93} & 92.23 & \textbf{82.06} & \textbf{98.73} \\
\hline
\multirow{3}{7em}{ResNet-50} & Real No Aug. & 82.81 & 82.84 & 82.81 & 81.41 & \textbf{92.38} & 52.18 & \textbf{96.72} \\
& Real with Aug. & 84.38 & 84.3 & 84.38 & 84.38 & 88.89 & 68.89 & 94.15 \\
& Real + Gen No Aug. & 86.02 & 85.8 & 85.94 & 85.16 & 92.00 & 66.91 & 96.55 \\
& Real + Gen with Aug. & \textbf{89.49} & \textbf{88.93} & \textbf{89.06} & \textbf{88.74} & 92.15 & \textbf{75.12} & 96.61 \\ \hline

\end{tabular}
}
\label{tab:results}
\end{table}

\begin{figure}[b!]
    \centering
    \includegraphics[width=1\linewidth]{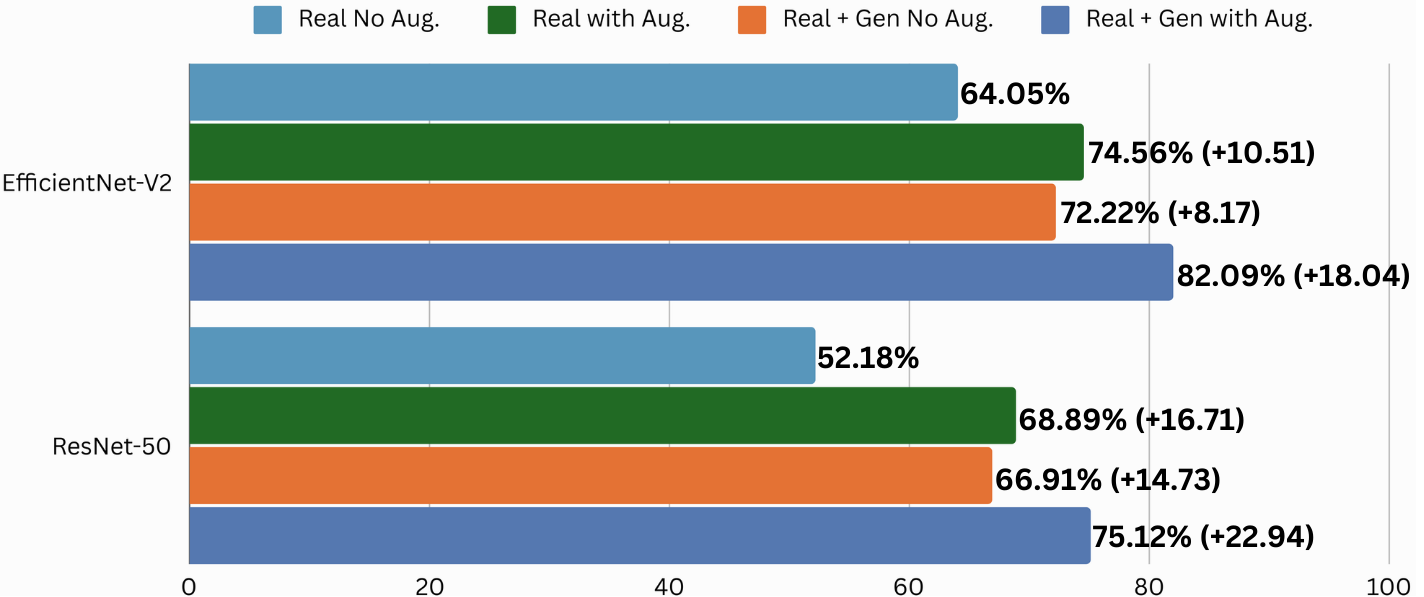}
    \caption{Comparison of Class 1 Accuracy Across Different Augmentation Techniques and Classifiers. Improvement points are with respect to the Real No Augmentation bar for each model.}
    \label{fig:barchart3}
\end{figure}

Notably, the most pronounced gains are observed in the accuracy metrics for the most challenging underrepresented class; class 1, where the conventional data pool is limited. The proposed strategy achieves the best performance in the majority of the experiments, being a very close second in the remaining cases.

The impact of different augmentation techniques on Class-1 accuracy is further highlighted in Figure \ref{fig:barchart3}. The results confirm that synthetic data augmentation, particularly when combined with traditional augmentation techniques, substantially improves the robustness of classification models.

EfficientNet-V2, when trained with both real and synthetically augmented data, shows an overall accuracy boost from 85.94\% to 92.19\%, and a significant increase in Class-1 accuracy from 64.05\% to 82.06\%. Similarly, ResNet-50's performance escalates from 82.81\% to 89.49\% in overall accuracy, with Class-1 accuracy rising from 52.18\% to 75.12\%. These enhancements suggest that synthetic data not only supplements the training set but also instills a better understanding of the feature space associated with each class.

Importantly, the augmented data appears to guide the model towards a more detailed comprehension of the subtle distinctions within the EGD image classes. This is critical for clinical applications where the differentiation between varying levels of cleanliness directly impacts the diagnostic process and subsequent patient care. Therefore the use of VAEs for data augmentation could suppose an advancement for medical imaging fields struggling with data constraints.

\begin{figure}[t!]
    \centering
    \includegraphics[width=\linewidth]{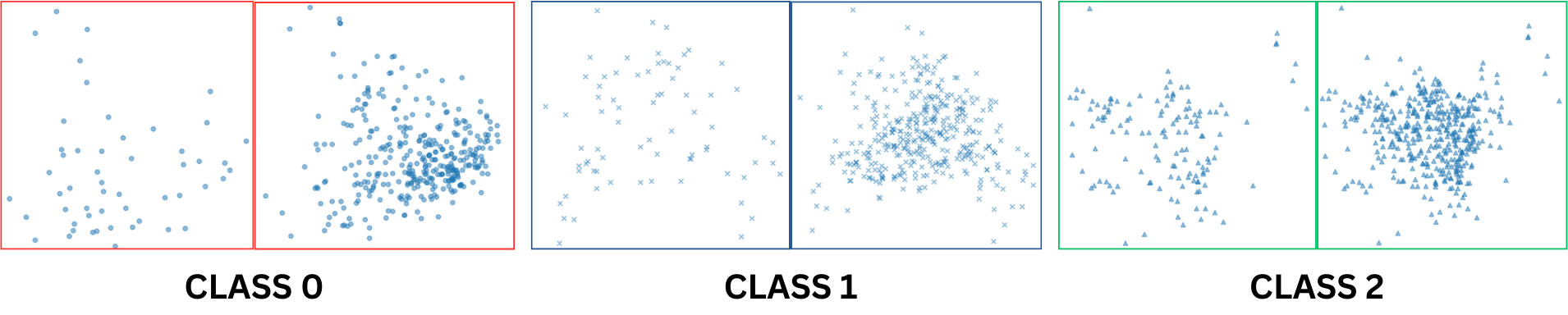}
    \caption{Expansion of feature space for each EGD image class post-augmentation. \textit{X} and \textit{Y} axes represent PCA features 1 and 2 respectively.}
    \label{fig:feature_space_expansion}
\end{figure}

Figure \ref{fig:feature_space_expansion} represents the data distribution for each class before and after synthetic augmentation. The original sparse distribution of each of the classes, as seen on the left side of each class's panel, becomes notably denser on the right side, following the application of VAE-based latent vector interpolation. This visual enhancement of the feature space is especially significant for Class 1, the primary focus of our study, where the augmented data points fill previously underpopulated regions, indicating a more balanced representation post-augmentation.

These visual findings align with the quantitative improvements in classification performance, confirming the value of VAE-based synthetic data in addressing class imbalance and enhancing model training for medical diagnostics.

\section{Conclusions and Future Work}

The research proposed in this paper demonstrates the effectiveness of synthetic data augmentation using class-specific Variational AutoEncoders (VAEs) for medical image classification, specifically targeting esophagogastroduodenoscopy (EGD) images. This is the first time such a system with these characteristics has been applied to the EGD imaging field, marking a significant impact in this domain. By interpolating latent representations within classes, a new method that significantly counters the limitations posed by small and unbalanced datasets has been developed and validated.

This approach has improved performance, particularly for challenging underrepresented classes, by effectively filling feature space gaps and achieving a more uniform dataset distribution. The success of this approach is demonstrated across two distinct architectures, EfficientNet-V2 and ResNet-50, showing its adaptability and the broad applicability of synthetic data augmentation in improving model classification capabilities.

Furthermore, the study explored the compounded benefits of combining traditional augmentation techniques with synthetic data augmentation, revealing a notable enhancement in the models' ability to generalize and accurately classify EGD images, discovering a synergistic effect.

The proposed work represents a significant step forward in utilizing AI for medical diagnostics, particularly by employing a methodologically innovative approach to synthetic data generation. By focusing on class-specific latent representation interpolations, it provides a scalable solution to the persisting problem of data scarcity and imbalance in medical imaging. This innovative methodology has set a precedent in the EGD imaging field, paving the way for its potential application in other medical imaging domains.

Despite these promising results, several limitations should be noted. The synthetic images generated by VAEs, while effective, may not capture all the nuances of real medical images, potentially leading to some biases in the training process. Additionally, the study was conducted on a relatively small dataset, which may limit the generalizability of the findings. 

This contribution lays the groundwork for future explorations into more sophisticated synthetic data generation methods and their application across various domains within medical image analysis. A compelling direction for this research could involve adopting latent diffusion models (LDMs), well known for their capacity to generate high-quality, realistic images, to augment the diversity and authenticity of synthetic medical images. This approach, coupled with assessing the impact of such augmentation techniques on larger and more diverse medical image datasets, could significantly advance the scalability, robustness and applicability of these methods.

Finally, refining the interpolation techniques for synthetic image creation to achieve more precise and clinically relevant datasets remains a critical area for development. Future efforts could also focus on understanding how synthetic data influences model interpretability and reliability in real-world clinical environments, aiming to not only elevate classification accuracy but also improve the trust and efficacy of diagnostic models in medical practice.


%
%
\begin{credits}
\subsubsection{\discintname} The authors have no competing interests to declare.
\end{credits}

\section*{Acknowledgements}

This work was supported by the following Grant Numbers: PID2020-120311RB-I00 and RED2022-134964-T and funded by MCIN-AEI/10.13039/501100011033.

\end{document}